\documentclass[11pt,a4paper]{article}
\usepackage[hyperref]{acl2021}
\usepackage{times}
\usepackage{latexsym}

\usepackage{microtype}

\aclfinalcopy 

\usepackage{url}

\usepackage{longtable}
\usepackage{tabu}
\usepackage{arydshln}

\usepackage{graphicx}
\usepackage{wrapfig}
\usepackage{CJK,algorithm,algorithmic,amssymb,amsmath,array,epsfig,graphics,multirow,array,float,subfigure,verbatim,epstopdf}
\usepackage{enumitem}
\usepackage{color,soul}
\usepackage{hhline}
\usepackage{multirow}
\usepackage{xcolor}
\usepackage{hyperref}
\usepackage{tabularx}
\usepackage{bm}
\usepackage{seqsplit}
\usepackage{stmaryrd}
\usepackage{booktabs}

\usepackage{xparse}

\usepackage{makecell}
\NewDocumentCommand{\vic}{ mO{} }{\textcolor{purple}{\textsuperscript{\textit{Victoria}}\textsf{\textbf{\small[#1]}}}}
\NewDocumentCommand{\semih}{ mO{} }{\textcolor{orange}{\textsuperscript{\textit{Semih}}\textsf{\textbf{\small[#1]}}}}
\NewDocumentCommand{\qingyun}{ mO{} }{\textcolor{blue}{\textsuperscript{\textit{Qingyun}}\textsf{\textbf{\small[#1]}}}}

\NewDocumentCommand{\heng}{ mO{} }{\textcolor{red}{\textsuperscript{\textit{Heng}}\textsf{\textbf{\small[#1]}}}}

\title{Stage-wise Fine-tuning for Graph-to-Text Generation}

\author{
Qingyun Wang$^{1}$\thanks{\ \ This research was conducted during the author’s internship at Salesforce Research.}, \ Semih Yavuz$^2$, \ Xi Victoria Lin$^{3}$, \\ \  \ \textbf{Heng Ji}$^{1}$, \  \textbf{Nazneen Fatema Rajani}$^2$\\   
$^{1}$ University of Illinois at Urbana-Champaign$^{2}$ Salesforce Research  $^{3}$ Facebook AI \\
{\tt syavuz,nazneen.rajani@salesforce.com}\\
  {\tt victorialin@fb.com}\\
  \texttt{\fontfamily{pcr}\selectfont\{qingyun4,hengji\}@illinois.edu}\\
}

\definecolor{fl1}{RGB}{221,223,0}
\definecolor{fl4}{RGB}{157,78,221}
\definecolor{fl5}{RGB}{191,210,0}
\definecolor{fl6}{RGB}{238,239,32}
\definecolor{fl7}{RGB}{128,185,24}
\definecolor{fl9}{RGB}{186,255,201}
\definecolor{fl10}{RGB}{186,225,25}
\definecolor{bleudefrance}{rgb}{0.19, 0.55, 0.91}
\definecolor{auburn}{rgb}{0.43, 0.21, 0.1}
\definecolor{ao(english)}{rgb}{0.0, 0.5, 0.0}
\date{}

\begin{document}

\maketitle

\begin{abstract}
Graph-to-text generation has benefited from pre-trained language models (PLMs) in achieving better performance than structured graph encoders. However, they fail to fully utilize the structure information of the input graph. 
In this paper, we aim to further improve the performance of the pre-trained language model by proposing 
a structured graph-to-text model with a two-step fine-tuning mechanism which first fine-tunes the model on Wikipedia before adapting to the graph-to-text generation. 
In addition to using the traditional token and position embeddings to encode the knowledge graph (KG), we propose a novel tree-level embedding method to capture the inter-dependency structures of the input graph. This new approach has significantly improved the performance of all text generation metrics for the English WebNLG 2017 dataset.\footnote{The programs, data and resources are publicly available for research purpose at: \url{https://github.com/EagleW/Stage-wise-Fine-tuning}}

\end{abstract}

\section{Introduction}

\begin{table*}[ht!]
\small
\begin{tabularx}{\linewidth}{>{\hsize=.15\hsize}X>{\hsize=1.85\hsize}X}
\toprule
\textbf{Category}&\textbf{Output}\\
\midrule
Reference&The Acharya Institute of Technology in \colorbox{fl1!50}{Karnataka} state was given Technical Campus status by \colorbox{fl7!50}{All} \colorbox{fl7!50}{India Council for Technical Education} in \colorbox{fl10!50}{Mumbai}. The school offers \colorbox{fl4!50}{tennis} which is governed by the \colorbox{fl9!50}{International Tennis Federation}. Karnataka has the \colorbox{fl5!50}{Arabian Sea} to its west and in the northeast is \colorbox{fl6!50}{Telangana}.
\\\hdashline
T5-large&The state of \colorbox{fl1!50}{Karnataka} is located southwest of \colorbox{fl6!50}{Telangana} and east of the \colorbox{fl5!50}{Arabian Sea}. It is the location of the Acharya Institute of Technology which was granted the Technical Campus status by the \colorbox{fl7!50}{All India Council for Technical Education} in \colorbox{fl10!50}{Mumbai}. The Institute is affiliated with the \textcolor{bleudefrance}{\textit{Visvesvaraya Technological University}} and offers the sport of \colorbox{fl4!50}{tennis}. \textcolor{ao(english)}{[International Tennis Federation]}
\\\hdashline
T5-large \newline+ Wiki& The Acharya Institute of Technology is located in the state of \colorbox{fl1!50}{Karnataka}. It was given the Technical Campus status by the \colorbox{fl7!50}{All India Council for Technical Education} which is located in \colorbox{fl10!50}{Mumbai}. The institute offers \colorbox{fl4!50}{tennis} and \textcolor{auburn}{\textbf{has \colorbox{fl6!50}{Telangana} to its northeast and the \colorbox{fl5!50}{Arabian Sea} to its west}}. \textcolor{ao(english)}{[International Tennis Federation]}
\\\hdashline
T5-large +\newline Position &The Acharya Institute of Technology is located in the state of \colorbox{fl1!50}{Karnataka} which has \colorbox{fl6!50}{Telangana} to its northeast and the \colorbox{fl5!50}{Arabian Sea} to its west. It was given the Technical Campus status by the \colorbox{fl7!50}{All India Council for Technical Education} in \colorbox{fl10!50}{Mumbai}. The Institute offers \colorbox{fl4!50}{tennis} which is governed by the \colorbox{fl9!50}{International Tennis Federation}.
\\\hdashline
T5-large \newline+ Wiki + \newline Position & 
The Acharya Institute of Technology in \colorbox{fl1!50}{Karnataka} was given the 'Technical Campus' status by the \colorbox{fl7!50}{All India Council for Technical Education} in \colorbox{fl10!50}{Mumbai}. Karnataka has \colorbox{fl6!50}{Telangana} to its northeast and the \colorbox{fl5!50}{Arabian Sea} to its west. One of the sports offered at the Institute is \colorbox{fl4!50}{tennis} which is governed by the \colorbox{fl9!50}{International Tennis Federation}.\\
\bottomrule
\end{tabularx}
\caption{\label{table:walkthrough}Human and System Generated Description in Figure \ref{fig:KG}. We use the color box to frame each entity out with the same color as the corresponding entity in Figure \ref{fig:KG}. We highlight \textcolor{bleudefrance}{\textit{fabricated facts}}, \textcolor{ao(english)}{[missed relations]}, and \textcolor{auburn}{\textbf{incorrect relations}} with different color. }

\end{table*}

\begin{figure}[!t]
\centering
\includegraphics[width=\linewidth]{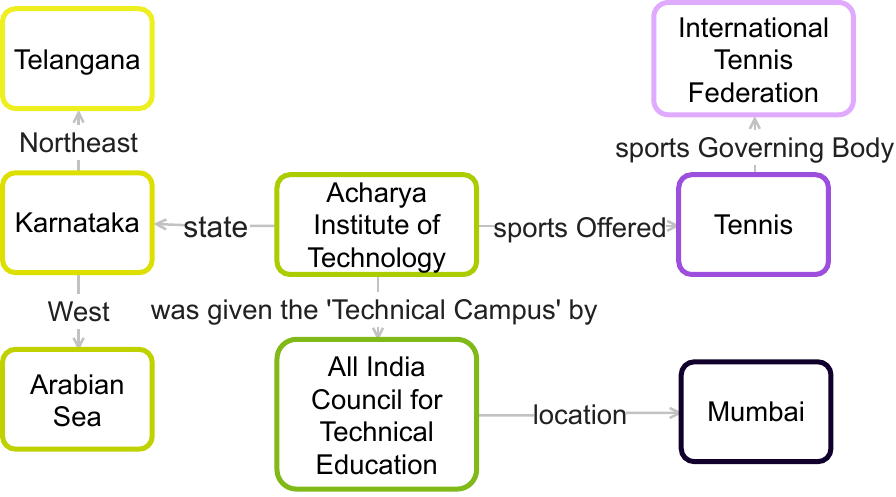}
\caption{\label{fig:KG}Input RDF Knowledge Graph
}
\end{figure}

In the graph-to-text generation task \cite{gardent-etal-2017-webnlg}, the model takes in a complex KG (an example is in Figure \ref{fig:KG}) and generates a corresponding faithful natural language description (Table \ref{table:walkthrough}). Previous efforts 
for this task 
can be mainly divided into two categories: sequence-to-sequence models that directly solve the generation task with LSTMs \cite{gardent-etal-2017-webnlg} or Transformer \cite{castro-ferreira-etal-2019-neural}; and graph-to-text models  \cite{trisedya-etal-2018-gtr,marcheggiani-perez-beltrachini-2018-deep} which use a graph encoder to capture the structure of the KGs. Recently, Transformer-based PLMs such as GPT-2~\cite{radford2019language}, BART~\cite{lewis-etal-2020-bart}, and T5~\cite{JMLR:v21:20-074} have achieved state-of-the-art results on WebNLG dataset due to factual knowledge acquired 
in the pre-training phase \cite{harkous-etal-2020-text,ribeiro2020investigating,kale2020text,chen-etal-2020-kgpt}.

Despite such improvement, PLMs fine-tuned only on the clean (or labeled) data might be more prone to hallucinate factual knowledge (e.g.,  \textit{``Visvesvaraya Technological University''} in Table \ref{table:walkthrough}).
Inspired by the success of domain-adaptive pre-training \cite{gururangan-etal-2020-dont}, we propose a novel two-step fine-tuning mechanism graph-to-text generation task.
Unlike \cite{ribeiro2020investigating,herzig-etal-2020-tapas,chen-etal-2020-kgpt} which directly fine-tune the PLMs on the training set, we first fine-tune our model over noisy RDF graphs and related article pairs crawled from Wikipedia before final fine-tuning on the clean/labeled training set. The additional fine-tuning step benefits our model by leveraging triples not included in the training set and reducing the chances that the model fabricates facts based on the language model.

Meanwhile, the PLMs might also fail to cover all relations in the KG by creating incorrect or missing facts. For example, in Table \ref{table:walkthrough}, 
although the T5-large with Wikipedia fine-tuning successfully removes the unwanted contents, 
it still ignores the \textit{``sports Governing Body''} relation and incorrectly links the university to both \textit{``Telangana''} and \textit{``Arabian Sea''}. To better capture the structure and interdependence of facts in the KG, instead of using a complex graph encoder, we leverage the power of Transformer-based PLMs with additional position embeddings which have been proved effective in various generation tasks \cite{herzig-etal-2020-tapas,chen-etal-2020-kgpt,Chen2020TabFact:}. Here, we extend the embedding layer of Transfomer-based PLMs with two additional \textit{triple role} and \textit{tree-level} embeddings to capture graph structure.

We explore the proposed stage-wise fine-tuning and structure-preserving embedding strategies for graph-to-text generation task on WebNLG corpus \cite{gardent-etal-2017-webnlg}. Our experimental results clearly demonstrate the benefit of each strategy in achieving the state-of-the-art performance on most commonly reported automatic evaluation metrics.

\section{Method}

  \begin{table*}[!htb]
\centering
\small
\begin{tabularx}{\linewidth}{>{\centering\arraybackslash\hsize=1.5\hsize}X>{\arraybackslash\hsize=3.5\hsize}X>{\centering\arraybackslash\hsize=0.5\hsize}X>{\centering\arraybackslash\hsize=0.9\hsize}X>{\centering\arraybackslash\hsize=0.5\hsize}X>{\centering\arraybackslash\hsize=0.5\hsize}X>{\centering\arraybackslash\hsize=0.9\hsize}X>{\centering\arraybackslash\hsize=0.5\hsize}X>{\centering\arraybackslash\hsize=0.5\hsize}X>{\centering\arraybackslash\hsize=0.9\hsize}X>{\centering\arraybackslash\hsize=0.5\hsize}X}
\toprule
    & \multirow{2}{*}{\textbf{Model}}&    \multicolumn{3}{c}{\textbf{BLEU}(\%)$\uparrow$} &  \multicolumn{3}{c}{\textbf{METEOR}$\uparrow$}&  \multicolumn{3}{c}{\textbf{TER}$\downarrow$}\\
\cline{3-11}
  & &    Seen &Unseen &All &    Seen &Unseen &All &    Seen &Unseen &All\\
\midrule
Without& \citet{gardent-etal-2017-webnlg}
             &54.52 & 33.27 & 45.13 & 0.41 & 0.33 & 0.37 & 0.40 & 0.55 & 0.47\\ 
Pretrained & \citet{moryossef-etal-2019-step} \footnotemark
             &53.30 & 34.41 & 47.24 & 0.44 & 0.34 & 0.39 & 0.47 & 0.56 & 0.51\\ 
 LM & \citet{zhao-etal-2020-bridging}
             &64.42 & 38.23 & 52.78 & 0.45 & 0.37 & 0.41 & 0.33 & 0.53 & 0.42\\ 
\hdashline
With   &
\citet{nan-etal-2021-dart} & 52.86 & 37.85 &45.89& 0.42& 0.37 &0.40& 0.44 &0.59 &0.51\\
Pretrained&\citet{kale2020text}&63.90 & 52.80 & 57.10 & \textbf{0.46} & 0.41 & \textbf{0.44} & -    & -    & -\\
LM&\citet{ribeiro2020investigating}&64.71 & 53.67 & 59.70 & \textbf{0.46} & \textbf{0.42} & \textbf{0.44} & -    & -    & -   \\
\hdashline
Our model&T5-large + Wiki + Position &\textbf{66.07} & \textbf{53.87} & \textbf{60.56} & \textbf{0.46} & \textbf{0.42} & \textbf{0.44} & \textbf{0.32} & \textbf{0.41} & \textbf{0.36}\\

\bottomrule
\end{tabularx}

\caption{\label{tab:webnlg}System Results on WebNLG Test Set Evaluated by BLEU, METEOR, and TER with Official Scripts
} 
\end{table*} 
Given an RDF graph with multiple relations $G=\{(s_1,r_1,o_1),(s_2,r_2,o_2),$ $...,(s_n,r_n,o_n)\}$,
our goal is to generate a text faithfully describing the input graph.
We represent each relation with a triple $(s_i,r_i,o_i) \in G$ for $i\in \{1,...,n\}$, where $s_i$, $r_i$, and $o_i$ are natural language phrases that represent the subject, type, and object of the relation, respectively.
\begin{figure}[!hbt]
\centering
\includegraphics[width=\linewidth]{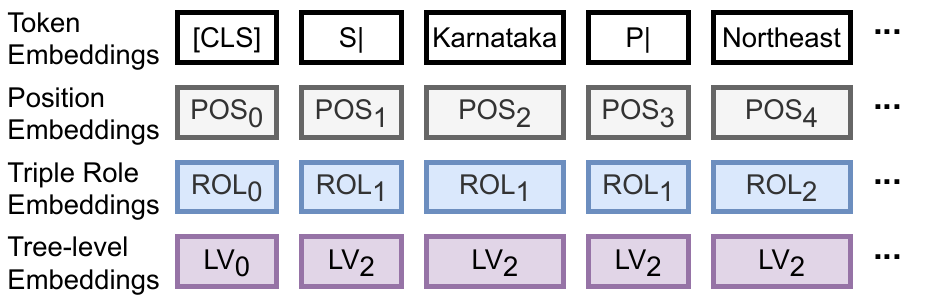}
\caption{\label{fig:generation}Position Embeddings for the KG in Figure \ref{fig:KG}
}
\end{figure}
We augment our model with additional position embeddings to capture the structure of the KG. To feed the input for the large-scale Transformer-based PLM, we flatten the graph as a concatenation of linearized triple sequences: 
\begin{align*}
    |S \ s_1 \ |P \ r_1 \ |O \ o_1 \ ... \ |S \ s_n \ |P \ r_n \ |O \ o_n
\end{align*}
following \citet{ribeiro2020investigating}, where $|S,|P,|O$ are special tokens prepended to indicate whether the phrases in the relations are subjects, relations, or objects, respectively.
Instead of directly fine-tuning the PLM on the WebNLG dataset, we first fine-tune our model on a noisy, but larger corpus crawled from Wikipedia, then we fine-tune the model on the training set.

\noindent\textbf{Positional embeddings} Since the input of the WebNLG task is a small KG which describes properties of entities, we introduce additional positional embeddings to enhance the flattened input of pre-trained Transformer-based sequence-to-sequence models such as BART and TaPas~\cite{herzig-etal-2020-tapas}.
\footnotetext{For this baseline, we use the results  reported from \citet{zhao-etal-2020-bridging} who also use official evaluation scripts.}We extend the input layer with two position-aware embeddings in addition to the original position embeddings\footnote{For T5 models, we only keep the Triple Role and Tree-level embeddings.} as shown in the Figure \ref{fig:generation}:
\begin{itemize}
    \item Position ID, which is the same as the original position ID used in BART, is the index of the token in the flattened sequence $|S$ $s_1$ $|P$ $r_1$ $|O$ $o_1$ ... $|S$ $s_n$ $|P$ $r_n$ $|O$ $o_n$ .
    \item Triple Role ID takes 3 values for a specific triple $(s_i,r_i,o_i)$: 1 for the subject $s_i$, 2 for the relation $r_i$, and 3 for the object $o_i$.
    \item Tree level ID calculates the distance (the number of relations) from the root which is the source vertex of the RDF graph.
\end{itemize}

\noindent\textbf{Two-step Fine-tuning} To get better domain adaptation ability \cite{gururangan-etal-2020-dont,herzig-etal-2020-tapas}, following TaPas and Wikipedia Person and Animal Dataset \cite{wang-etal-2018-describing}, we perform intermediate pre-training by coupling noisy English Wikipedia data with Wikidata triples, both of which are crawled in March 2020. 
We select 15 related categories (Astronaut,
University, Monument, Building, ComicsCharacter, Food, Airport, SportsTeam, WrittenWork, Athlete, Artist, City, MeanOfTransportation, CelestialBody, Politician) that appear in the WebNLG dataset \cite{gardent-etal-2017-webnlg} and collect 542,192 data pairs. For each Wikipedia article, we query its corresponding WikiData triples and remove sentences which contain no values in the Wikidata triples to form graph-text pairs. Unlike \cite{chen-etal-2020-kgpt} which focuses on individual entity-sentence pairs for distant supervision, our pre-training corpus, on the other hand, is designed to better adapt to translating deeper graph structure into text.
We remove triples and description pairs that have already appeared in the WebNLG dataset. After intermediate pre-training on this noisy corpus, we continue with fine-tuning our model on the WebNLG dataset.

\section{Experiments}

\subsection{Dataset and Implementation details}
\begin{table}[!htb]
\centering
\small
\begin{tabularx}{\linewidth}{>{\hsize=2.5\hsize}X>{\centering\arraybackslash\hsize=1\hsize}X>{\centering\arraybackslash\hsize=0.5\hsize}X>{\centering\arraybackslash\hsize=0.5\hsize}X>{\centering\arraybackslash\hsize=0.5\hsize}X}
\toprule
\textbf{Model}&\textbf{BLEU}$\uparrow$&\textbf{P}$\uparrow$&\textbf{R}$\uparrow$&\textbf{F1}$\uparrow$\\
\midrule
BART-base               & 57.8& 68.7& 68.9& 67.0\\
\hspace{2mm}+ Wikipedia & \textbf{59.7}&\textbf{69.6} & \textbf{70.7}& \textbf{68.4}\\
\hspace{2mm}+ Position  & 58.8& 68.7& 69.9& 67.6\\
\hspace{2mm}+ Wiki + Position & 57.3& 67.8& 69.0& 66.6\\
\hdashline
BART-large              & 58.3& 67.9& 69.4& 66.8\\
\hspace{2mm}+ Wikipedia & 59.0& 68.0& \textbf{70.4}& \textbf{67.4}\\
\hspace{2mm}+ Position  & 58.1& 67.6& 69.4& 66.6\\
\hspace{2mm}+ Wiki + Position & \textbf{60.0}& \textbf{68.6}& 69.2& 67.1\\
\hdashline
distill-BART-xsum       & 59.1& 69.9& 70.6& 68.5\\
\hspace{2mm}+ Wikipedia & 59.8& 69.7& \textbf{71.1}& \textbf{68.8}\\
\hspace{2mm}+ Position  & 59.2& 69.8& 70.2& 68.3\\
\hspace{2mm}+ Wiki + Position & \textbf{59.9}& \textbf{70.1}& 70.1& 68.7\\
\hdashline
T5-base                 & \textbf{61.2}& 72.3& 72.0& 70.6\\
\hspace{2mm}+ Wikipedia & 60.9& 72.0& 71.8& 70.2\\
\hspace{2mm}+ Position  & 60.8& \textbf{72.4}& \textbf{72.4}& \textbf{70.8}\\
\hspace{2mm}+ Wiki + Position & 60.3& 72.2& 72.0& 70.5\\
\hdashline
T5-large                & 60.0& 71.6& 72.1& 70.2\\
\hspace{2mm}+ Wikipedia & 61.3& 72.2& 72.0& 70.5\\
\hspace{2mm}+ Position  & 60.6& 72.1& 72.4& 70.6\\
\hspace{2mm}+ Wiki + Position & \textbf{61.9}& \textbf{72.8}& \textbf{73.5}& \textbf{71.6}\\
\bottomrule
\end{tabularx}
\caption{Results with both Wikipedia Fine-tuning and Positional Embedding for Various Pre-trained Models over All Categories on Development Set Evaluated by average of PARENT\footnotemark precision, recall, F1 and BLEU (\%) \label{tab:dev} }
\end{table}
\footnotetext{\url{https://github.com/KaijuML/parent}}

We use the original version of English WebNLG2017 \cite{gardent-etal-2017-webnlg} dataset which contains 18,102/2,268/4,928
graph-description pairs for training, validation, and testing set respectfully. For this task, we investigate a variety of the BART and T5 models with our novel tree-level embeddings. The statistics and more details of those models are listed in Appendix A.

\begin{table}[!htb]
\centering
\small
\begin{tabularx}{\linewidth}{>{\hsize=2.5\hsize}X>{\centering\arraybackslash\hsize=0.5\hsize}X>{\centering\arraybackslash\hsize=0.5\hsize}X>{\centering\arraybackslash\hsize=0.5\hsize}X}
\toprule
\textbf{Model}&\textbf{P}$\uparrow$&\textbf{R}$\uparrow$&\textbf{F1}$\uparrow$\\
\midrule
\citet{gardent-etal-2017-webnlg}
             &88.35 & 90.22 & 89.23 \\ 
\citet{moryossef-etal-2019-step} 
             &85.77 & 89.34 & 87.46 \\ 
\hdashline
\citet{nan-etal-2021-dart} & 89.49 & 92.33 &90.83\\
\citet{ribeiro2020investigating}&89.36 & 91.96 & 90.59 \\
\hdashline
T5-large + Wiki + Position &\textbf{96.36}&\textbf{96.13}&\textbf{96.21}\\
\bottomrule
\end{tabularx}
\caption{System Results on WebNLG Test Set Evaluated by BERTScore
precision, recall, F1 (\%)\label{tab:bert} }
\end{table}
\begin{table*}[ht!]
\small
\begin{tabularx}{\linewidth}{>{\hsize=.15\hsize}X>{\hsize=1.85\hsize}X}
\toprule
\textbf{Category}&\textbf{Output}\\
\midrule
T5-large&Andrew White \textcolor{bleudefrance}{\textit{(born in 2003)}} is a musician who is associated with the band Kaiser Chiefs and Marry Banilow. He is also associated with the label Polydor Records and is signed to B-Unique Records. \textcolor{red}{{S$|$ Aleksandra Kovač P$|$ activeYearsStartYear O$|$ 1990}}\\
T5-large&Walter Baade was born in the German Empire and graduated from the University of Gottingen.  \textcolor{auburn}{\textbf{He was the doctoral student of Halton Arp and Allan Sandage}} and was the discoverer of 1036 Ganymed.
\textcolor{red}{ S$|$ Walter Baade P$|$ doctoralStudent O$|$ Halton Arp; S$|$ Walter Baade P$|$ doctoralStudent O$|$ Allan Sandage} \\
T5-large\newline+Wiki&11264 Claudiomaccone was \textcolor{bleudefrance}{\textit{born on the 26th of November, 2005}}. He has an orbital period of 1513.722 days, a periapsis of 296521000.0 kilometres and an apoapsis of 475426000.0 kilometres. \textcolor{red}{ S$|$ 11264 Claudiomaccone P$|$ epoch O$|$ 2005-11-26; S$|$ Aleksandr Prudnikov P$|$ club O$|$ FC Amkar Perm  }\\
T5-large\newline+Position&The chairman of FC Spartak Moscow is Sergey Rodionov. Aleksandr Prudnikov plays for FC Spartak Moscow and \textcolor{bleudefrance}{\textit{manages FC Amkar Perm}}. \textcolor{ao(english)}{[ S$|$ FC Amkar Perm P$|$ manager O$|$ Gadzhi Gadzhiyev; S$|$ Aleksandr Prudnikov P$|$ club O$|$ FC Amkar Perm ]}\\
\bottomrule
\end{tabularx}
\caption{System Error Examples. We highlight \textcolor{bleudefrance}{\textit{ fabricated facts}}, \textcolor{ao(english)}{[missed relations]}, \textcolor{auburn}{\textbf{incorrect relations}}, and \textcolor{red}{ground truth relations} with different color. \label{tab:error}}

\end{table*}
\subsection{Results and Analysis}

We use the standard NLG evaluation metrics to report results: BLEU~\cite{papineni-etal-2002-bleu}, METEOR \cite{lavie-agarwal-2007-meteor}, and TER \cite{Snover06astudy}
, as shown in Table~\ref{tab:webnlg}. Because \citet{castro-ferreira-etal-2020-2020} has found that BERTScore~\cite{Zhang*2020BERTScore:} correlates with human evaluation ratings better, we use BERTscore to evaluate system results\footnote{We only use BERTScore to evaluate baselines which have results available online.} as shown in Table \ref{tab:bert}.  When selecting the best models, 
we also evaluate each model with PARENT~\cite{dhingra-etal-2019-handling} metric which measures the overlap between predictions and both reference texts and graph contents. \citet{dhingra-etal-2019-handling} show PARENT metric has better human rating correlations.
Table \ref{tab:dev} shows the pre-trained models with 2-step fine-tuning and position embeddings achieve better results.\footnote{For more examples, please check Appendix for reference.} We conduct paired t-test between our proposed model and all the other baselines on 10 randomly sampled subsets. The differences are statistically significant with $p \leq 0.008$ for all settings.

\paragraph{\textbf{Results with Wikipedia fine-tuning.}} The Wikipedia fine-tuning helps the model handle unseen relations such as \textit{``inOfficeWhileVicePresident''}, and \textit{``activeYearsStartYear''} by stating \textit{``\textbf{His vice president} is Atiku Abubakar.''} and \textit{``\textbf{started playing} in 1995''} respectively. It also combines relations with the same type together with correct order, e.g., given two death places of a person, the model generates: \textit{``died in \textbf{Sidcup, London}''} instead of generating two sentences or placing the city name ahead of the area name.

\paragraph{\textbf{Results with positional embeddings.}}For the KG with multiple triples, additional positional embeddings help reduce the errors introduced by pronoun ambiguity. For instance, for a KG which has \textit{``leaderName''} relation to both country's leader and university's dean, position embeddings can distinguish these two relations
by stating \textit{``\textbf{Denmark's} leader is Lars Løkke Rasmussen''} instead of \textit{``\textbf{its} leader is Lars Løkke Rasmussen''}. The tree-level embeddings also help the model arrange multiple triples into one sentence, such as combining the city, the country, the affiliation, and the affiliation's headquarter of a university into a single sentence: \textit{``The School of Business and Social Sciences at the Aarhus University in \textbf{Aarhus}, \textbf{Denmark} is affiliated to \textbf{the European University Association} in \textbf{Brussels}''}.

\subsection{Remaining Challenges}
However, pre-trained language models also generate some errors as shown in Table \ref{tab:error}.
Because the language model is heavily pre-trained, it is biased against the occurrence of patterns that would enable it to infer the right relation. For example, for the \textit{``activeYearsStartYear''}
 relation, the model might confuse it with the birth year. For some relations that do not have a clear direction, the language model is not powerful enough to consider the deep connections between the subject and the object. For example, for the relation \textit{``doctoralStudent''}, the model mistakenly describes a professor as a Ph.D. student. Similarly, the model treats an asteroid as a person because it has an epoch date. For KGs with multiple triples, the generator still has a chance to miss relations or mixes the subject and the object of different relations, especially for the unseen category. For instance, for a soccer player with multiple clubs, the system might confuse the subject of one club's relation with another club.

\section{Related Work}

The WebNLG task is similar to Wikibio generation \cite{lebret-etal-2016-neural,wang-etal-2018-describing}, AMR-to-text generation \cite{song-etal-2018-graph} and ROTOWIRE \cite{wiseman-etal-2017-challenges,puduppully2019data}. Previous methods usually treat the graph-to-text generation as an end-to-end generation task. Those models \cite{trisedya-etal-2018-gtr,gong-etal-2019-table,shen-etal-2020-neural} usually first lineralize the knowledge graph and then use attention mechanism to generate the description sentences.
While the linearization of input graph may sacrifice the inter-dependency inside input graph, some papers \cite{ribeiro-etal-2019-enhancing,doi:10.1162tacla00332,zhao-etal-2020-bridging} use graph encoder such as GCN \cite{NIPS2015_5954} and graph transformer \cite{wang-etal-2020-amr,koncel-kedziorski-etal-2019-text} to encode the input graphs. Others \cite{shen-etal-2020-neural,wang-etal-2020-towards} try to carefully design loss functions to control the generation quality. With the development of computation resources, large scale PLMs such as GPT-2 \cite{radford2019language}, BART \cite{lewis-etal-2020-bart} and T5~\cite{JMLR:v21:20-074} achieve state-of-the-art results even with simple linearized graph input \cite{harkous-etal-2020-text,chen-etal-2020-kgpt,kale2020text,ribeiro2020investigating}. Instead of directly fine-tuning the PLMs, we propose a two-step fine-tuning mechanism to get better domain adaptation ability. In addition, using positional embeddings as an extension for PLMs has shown its effectiveness in table-based question answering \cite{herzig-etal-2020-tapas}, fact verification \cite{Chen2020TabFact:}, and graph-to-text generation \cite{chen-etal-2020-kgpt}. We capture the graph structure by enhancing the input layer with the triple role and tree-level embeddings.

\section{Conclusions and Future Work}

We propose a new two-step structured generation task for the graph-to-text generation task based on a two-step fine-tuning mechanism and novel tree-level position embeddings. In the future, we aim to address the remaining challenges and extend the framework for broader applications.

\section*{Acknowledgement}
This work is partially supported by Agriculture and Food Research Initiative (AFRI) grant no. 2020-67021-32799/project accession no.1024178 from the USDA National Institute of Food and Agriculture, and by the Office of the Director of National Intelligence (ODNI), Intelligence Advanced Research Projects Activity (IARPA), via contract \# FA8650-17-C-9116. The views and conclusions contained herein are those of the authors and should not be interpreted as necessarily representing the official policies, either expressed or implied of the U.S. Government. The U.S. Government is authorized to reproduce and distribute reprints for governmental purposes notwithstanding any copyright annotation therein.

\bibliography{anthology,custom}

\begin{thebibliography}{38}
\expandafter\ifx\csname natexlab\endcsname\relax\def\natexlab#1{#1}\fi

\bibitem[{Castro~Ferreira et~al.(2020)Castro~Ferreira, Gardent, Ilinykh,
  van~der Lee, Mille, Moussallem, and
  Shimorina}]{castro-ferreira-etal-2020-2020}
Thiago Castro~Ferreira, Claire Gardent, Nikolai Ilinykh, Chris van~der Lee,
  Simon Mille, Diego Moussallem, and Anastasia Shimorina. 2020.
\newblock \href {https://www.aclweb.org/anthology/2020.webnlg-1.7} {The 2020
  bilingual, bi-directional {W}eb{NLG}+ shared task: Overview and evaluation
  results ({W}eb{NLG}+ 2020)}.
\newblock In \emph{Proceedings of the 3rd International Workshop on Natural
  Language Generation from the Semantic Web (WebNLG+)}, pages 55--76, Dublin,
  Ireland (Virtual). Association for Computational Linguistics.

\bibitem[{Castro~Ferreira et~al.(2019)Castro~Ferreira, van~der Lee, van
  Miltenburg, and Krahmer}]{castro-ferreira-etal-2019-neural}
Thiago Castro~Ferreira, Chris van~der Lee, Emiel van Miltenburg, and Emiel
  Krahmer. 2019.
\newblock \href {https://doi.org/10.18653/v1/D19-1052} {Neural data-to-text
  generation: A comparison between pipeline and end-to-end architectures}.
\newblock In \emph{Proceedings of the 2019 Conference on Empirical Methods in
  Natural Language Processing and the 9th International Joint Conference on
  Natural Language Processing (EMNLP-IJCNLP)}, pages 552--562, Hong Kong,
  China. Association for Computational Linguistics.

\bibitem[{Chen et~al.(2020{\natexlab{a}})Chen, Su, Yan, and
  Wang}]{chen-etal-2020-kgpt}
Wenhu Chen, Yu~Su, Xifeng Yan, and William~Yang Wang. 2020{\natexlab{a}}.
\newblock \href {https://doi.org/10.18653/v1/2020.emnlp-main.697} {{KGPT}:
  Knowledge-grounded pre-training for data-to-text generation}.
\newblock In \emph{Proceedings of the 2020 Conference on Empirical Methods in
  Natural Language Processing (EMNLP)}, pages 8635--8648, Online. Association
  for Computational Linguistics.

\bibitem[{Chen et~al.(2020{\natexlab{b}})Chen, Wang, Chen, Zhang, Wang, Li,
  Zhou, and Wang}]{Chen2020TabFact:}
Wenhu Chen, Hongmin Wang, Jianshu Chen, Yunkai Zhang, Hong Wang, Shiyang Li,
  Xiyou Zhou, and William~Yang Wang. 2020{\natexlab{b}}.
\newblock \href {https://openreview.net/forum?id=rkeJRhNYDH} {Tabfact: A
  large-scale dataset for table-based fact verification}.
\newblock In \emph{Proceedings of the 8th International Conference on Learning
  Representations}.

\bibitem[{Dhingra et~al.(2019)Dhingra, Faruqui, Parikh, Chang, Das, and
  Cohen}]{dhingra-etal-2019-handling}
Bhuwan Dhingra, Manaal Faruqui, Ankur Parikh, Ming-Wei Chang, Dipanjan Das, and
  William Cohen. 2019.
\newblock \href {https://doi.org/10.18653/v1/P19-1483} {Handling divergent
  reference texts when evaluating table-to-text generation}.
\newblock In \emph{Proceedings of the 57th Annual Meeting of the Association
  for Computational Linguistics}, pages 4884--4895, Florence, Italy.
  Association for Computational Linguistics.

\bibitem[{Duvenaud et~al.(2015)Duvenaud, Maclaurin, Iparraguirre, Bombarell,
  Hirzel, Aspuru-Guzik, and Adams}]{NIPS2015_5954}
David~K Duvenaud, Dougal Maclaurin, Jorge Iparraguirre, Rafael Bombarell,
  Timothy Hirzel, Alan Aspuru-Guzik, and Ryan~P Adams. 2015.
\newblock \href
  {http://papers.nips.cc/paper/5954-convolutional-networks-on-graphs-for-learning-molecular-fingerprints.pdf}
  {Convolutional networks on graphs for learning molecular fingerprints}.
\newblock In C.~Cortes, N.~D. Lawrence, D.~D. Lee, M.~Sugiyama, and R.~Garnett,
  editors, \emph{Advances in Neural Information Processing Systems 28}, pages
  2224--2232. Curran Associates, Inc.

\bibitem[{Gardent et~al.(2017)Gardent, Shimorina, Narayan, and
  Perez-Beltrachini}]{gardent-etal-2017-webnlg}
Claire Gardent, Anastasia Shimorina, Shashi Narayan, and Laura
  Perez-Beltrachini. 2017.
\newblock \href {https://doi.org/10.18653/v1/W17-3518} {The {W}eb{NLG}
  challenge: Generating text from {RDF} data}.
\newblock In \emph{Proceedings of the 10th International Conference on Natural
  Language Generation}, pages 124--133, Santiago de Compostela, Spain.
  Association for Computational Linguistics.

\bibitem[{Gong et~al.(2019)Gong, Feng, Qin, and Liu}]{gong-etal-2019-table}
Heng Gong, Xiaocheng Feng, Bing Qin, and Ting Liu. 2019.
\newblock \href {https://doi.org/10.18653/v1/D19-1310} {Table-to-text
  generation with effective hierarchical encoder on three dimensions (row,
  column and time)}.
\newblock In \emph{Proceedings of the 2019 Conference on Empirical Methods in
  Natural Language Processing and the 9th International Joint Conference on
  Natural Language Processing (EMNLP-IJCNLP)}, pages 3143--3152, Hong Kong,
  China. Association for Computational Linguistics.

\bibitem[{Gururangan et~al.(2020)Gururangan, Marasovi{\'c}, Swayamdipta, Lo,
  Beltagy, Downey, and Smith}]{gururangan-etal-2020-dont}
Suchin Gururangan, Ana Marasovi{\'c}, Swabha Swayamdipta, Kyle Lo, Iz~Beltagy,
  Doug Downey, and Noah~A. Smith. 2020.
\newblock \href {https://doi.org/10.18653/v1/2020.acl-main.740} {Don{'}t stop
  pretraining: Adapt language models to domains and tasks}.
\newblock In \emph{Proceedings of the 58th Annual Meeting of the Association
  for Computational Linguistics}, pages 8342--8360, Online. Association for
  Computational Linguistics.

\bibitem[{Harkous et~al.(2020)Harkous, Groves, and
  Saffari}]{harkous-etal-2020-text}
Hamza Harkous, Isabel Groves, and Amir Saffari. 2020.
\newblock \href {https://doi.org/10.18653/v1/2020.coling-main.218} {Have your
  text and use it too! end-to-end neural data-to-text generation with semantic
  fidelity}.
\newblock In \emph{Proceedings of the 28th International Conference on
  Computational Linguistics}, pages 2410--2424, Barcelona, Spain (Online).
  International Committee on Computational Linguistics.

\bibitem[{Herzig et~al.(2020)Herzig, Nowak, M{\"u}ller, Piccinno, and
  Eisenschlos}]{herzig-etal-2020-tapas}
Jonathan Herzig, Pawel~Krzysztof Nowak, Thomas M{\"u}ller, Francesco Piccinno,
  and Julian Eisenschlos. 2020.
\newblock \href {https://doi.org/10.18653/v1/2020.acl-main.398} {{T}a{P}as:
  Weakly supervised table parsing via pre-training}.
\newblock In \emph{Proceedings of the 58th Annual Meeting of the Association
  for Computational Linguistics}, pages 4320--4333, Online. Association for
  Computational Linguistics.

\bibitem[{Kale(2020)}]{kale2020text}
Mihir Kale. 2020.
\newblock \href {https://arxiv.org/pdf/2005.10433.pdf} {Text-to-text
  pre-training for data-to-text tasks}.
\newblock \emph{Computation and Language Repository}, arXiv:2005.10433.
\newblock Version 2.

\bibitem[{Kingma and Ba(2015)}]{kingma2014adam}
Diederik~P Kingma and Jimmy Ba. 2015.
\newblock \href {https://arxiv.org/pdf/1412.6980.pdf} {Adam: A method for
  stochastic optimization}.
\newblock In \emph{Proceedings of the 3rd International Conference on Learning
  Representations}.

\bibitem[{Koncel-Kedziorski et~al.(2019)Koncel-Kedziorski, Bekal, Luan, Lapata,
  and Hajishirzi}]{koncel-kedziorski-etal-2019-text}
Rik Koncel-Kedziorski, Dhanush Bekal, Yi~Luan, Mirella Lapata, and Hannaneh
  Hajishirzi. 2019.
\newblock \href {https://doi.org/10.18653/v1/N19-1238} {{T}ext {G}eneration
  from {K}nowledge {G}raphs with {G}raph {T}ransformers}.
\newblock In \emph{Proceedings of the 2019 Conference of the North {A}merican
  Chapter of the Association for Computational Linguistics: Human Language
  Technologies, Volume 1 (Long and Short Papers)}, pages 2284--2293,
  Minneapolis, Minnesota. Association for Computational Linguistics.

\bibitem[{Lavie and Agarwal(2007)}]{lavie-agarwal-2007-meteor}
Alon Lavie and Abhaya Agarwal. 2007.
\newblock \href {https://www.aclweb.org/anthology/W07-0734} {{METEOR}: An
  automatic metric for {MT} evaluation with high levels of correlation with
  human judgments}.
\newblock In \emph{Proceedings of the Second Workshop on Statistical Machine
  Translation}, pages 228--231, Prague, Czech Republic. Association for
  Computational Linguistics.

\bibitem[{Lebret et~al.(2016)Lebret, Grangier, and
  Auli}]{lebret-etal-2016-neural}
R{\'e}mi Lebret, David Grangier, and Michael Auli. 2016.
\newblock \href {https://doi.org/10.18653/v1/D16-1128} {Neural text generation
  from structured data with application to the biography domain}.
\newblock In \emph{Proceedings of the 2016 Conference on Empirical Methods in
  Natural Language Processing}, pages 1203--1213, Austin, Texas. Association
  for Computational Linguistics.

\bibitem[{Lewis et~al.(2020)Lewis, Liu, Goyal, Ghazvininejad, Mohamed, Levy,
  Stoyanov, and Zettlemoyer}]{lewis-etal-2020-bart}
Mike Lewis, Yinhan Liu, Naman Goyal, Marjan Ghazvininejad, Abdelrahman Mohamed,
  Omer Levy, Veselin Stoyanov, and Luke Zettlemoyer. 2020.
\newblock \href {https://doi.org/10.18653/v1/2020.acl-main.703} {{BART}:
  Denoising sequence-to-sequence pre-training for natural language generation,
  translation, and comprehension}.
\newblock In \emph{Proceedings of the 58th Annual Meeting of the Association
  for Computational Linguistics}, pages 7871--7880, Online. Association for
  Computational Linguistics.

\bibitem[{Marcheggiani and
  Perez-Beltrachini(2018)}]{marcheggiani-perez-beltrachini-2018-deep}
Diego Marcheggiani and Laura Perez-Beltrachini. 2018.
\newblock \href {https://doi.org/10.18653/v1/W18-6501} {Deep graph
  convolutional encoders for structured data to text generation}.
\newblock In \emph{Proceedings of the 11th International Conference on Natural
  Language Generation}, pages 1--9, Tilburg University, The Netherlands.
  Association for Computational Linguistics.

\bibitem[{Moryossef et~al.(2019)Moryossef, Goldberg, and
  Dagan}]{moryossef-etal-2019-step}
Amit Moryossef, Yoav Goldberg, and Ido Dagan. 2019.
\newblock \href {https://doi.org/10.18653/v1/N19-1236} {{S}tep-by-step:
  {S}eparating planning from realization in neural data-to-text generation}.
\newblock In \emph{Proceedings of the 2019 Conference of the North {A}merican
  Chapter of the Association for Computational Linguistics: Human Language
  Technologies, Volume 1 (Long and Short Papers)}, pages 2267--2277,
  Minneapolis, Minnesota. Association for Computational Linguistics.

\bibitem[{Nan et~al.(2021)Nan, Radev, Zhang, Rau, Sivaprasad, Hsieh, Tang,
  Vyas, Verma, Krishna, Liu, Irwanto, Pan, Rahman, Zaidi, Mutuma, Tarabar,
  Gupta, Yu, Tan, Lin, Xiong, Socher, and Rajani}]{nan-etal-2021-dart}
Linyong Nan, Dragomir Radev, Rui Zhang, Amrit Rau, Abhinand Sivaprasad,
  Chiachun Hsieh, Xiangru Tang, Aadit Vyas, Neha Verma, Pranav Krishna,
  Yangxiaokang Liu, Nadia Irwanto, Jessica Pan, Faiaz Rahman, Ahmad Zaidi,
  Mutethia Mutuma, Yasin Tarabar, Ankit Gupta, Tao Yu, Yi~Chern Tan,
  Xi~Victoria Lin, Caiming Xiong, Richard Socher, and Nazneen~Fatema Rajani.
  2021.
\newblock \href {https://www.aclweb.org/anthology/2021.naacl-main.37} {{DART}:
  Open-domain structured data record to text generation}.
\newblock In \emph{Proceedings of the 2021 Conference of the North American
  Chapter of the Association for Computational Linguistics: Human Language
  Technologies}, pages 432--447, Online. Association for Computational
  Linguistics.

\bibitem[{Papineni et~al.(2002)Papineni, Roukos, Ward, and
  Zhu}]{papineni-etal-2002-bleu}
Kishore Papineni, Salim Roukos, Todd Ward, and Wei-Jing Zhu. 2002.
\newblock \href {https://doi.org/10.3115/1073083.1073135} {{B}leu: a method for
  automatic evaluation of machine translation}.
\newblock In \emph{Proceedings of the 40th Annual Meeting of the Association
  for Computational Linguistics}, pages 311--318, Philadelphia, Pennsylvania,
  USA. Association for Computational Linguistics.

\bibitem[{Puduppully et~al.(2019)Puduppully, Dong, and
  Lapata}]{puduppully2019data}
Ratish Puduppully, Li~Dong, and Mirella Lapata. 2019.
\newblock \href {https://ojs.aaai.org//index.php/AAAI/article/view/4668}
  {Data-to-text generation with content selection and planning}.
\newblock In \emph{Proceedings of the AAAI Conference on Artificial
  Intelligence}, volume~33, pages 6908--6915.

\bibitem[{Radford et~al.(2019)Radford, Wu, Child, Luan, Amodei, and
  Sutskever}]{radford2019language}
Alec Radford, Jeffrey Wu, Rewon Child, David Luan, Dario Amodei, and Ilya
  Sutskever. 2019.
\newblock \href
  {https://d4mucfpksywv.cloudfront.net/better-language-models/language-models.pdf}
  {Language models are unsupervised multitask learners}.
\newblock \emph{OpenAI blog}, 1(8):9.

\bibitem[{Raffel et~al.(2020)Raffel, Shazeer, Roberts, Lee, Narang, Matena,
  Zhou, Li, and Liu}]{JMLR:v21:20-074}
Colin Raffel, Noam Shazeer, Adam Roberts, Katherine Lee, Sharan Narang, Michael
  Matena, Yanqi Zhou, Wei Li, and Peter~J. Liu. 2020.
\newblock \href {http://jmlr.org/papers/v21/20-074.html} {Exploring the limits
  of transfer learning with a unified text-to-text transformer}.
\newblock \emph{Journal of Machine Learning Research}, 21(140):1--67.

\bibitem[{Ribeiro et~al.(2019)Ribeiro, Gardent, and
  Gurevych}]{ribeiro-etal-2019-enhancing}
Leonardo F.~R. Ribeiro, Claire Gardent, and Iryna Gurevych. 2019.
\newblock \href {https://doi.org/10.18653/v1/D19-1314} {Enhancing {AMR}-to-text
  generation with dual graph representations}.
\newblock In \emph{Proceedings of the 2019 Conference on Empirical Methods in
  Natural Language Processing and the 9th International Joint Conference on
  Natural Language Processing (EMNLP-IJCNLP)}, pages 3183--3194, Hong Kong,
  China. Association for Computational Linguistics.

\bibitem[{Ribeiro et~al.(2020{\natexlab{a}})Ribeiro, Zhang, Gardent, and
  Gurevych}]{doi:10.1162tacla00332}
Leonardo F.~R. Ribeiro, Yue Zhang, Claire Gardent, and Iryna Gurevych.
  2020{\natexlab{a}}.
\newblock \href {https://doi.org/10.1162/tacl\_a\_00332} {Modeling global and
  local node contexts for text generation from knowledge graphs}.
\newblock \emph{Transactions of the Association for Computational Linguistics},
  8:589--604.

\bibitem[{Ribeiro et~al.(2020{\natexlab{b}})Ribeiro, Schmitt, Sch{\"u}tze, and
  Gurevych}]{ribeiro2020investigating}
Leonardo~FR Ribeiro, Martin Schmitt, Hinrich Sch{\"u}tze, and Iryna Gurevych.
  2020{\natexlab{b}}.
\newblock \href {https://arxiv.org/pdf/2007.08426.pdf} {Investigating
  pretrained language models for graph-to-text generation}.
\newblock \emph{arXiv preprint arXiv:2007.08426}.

\bibitem[{Shen et~al.(2020)Shen, Chang, Su, Niu, and
  Klakow}]{shen-etal-2020-neural}
Xiaoyu Shen, Ernie Chang, Hui Su, Cheng Niu, and Dietrich Klakow. 2020.
\newblock \href {https://doi.org/10.18653/v1/2020.acl-main.641} {Neural
  data-to-text generation via jointly learning the segmentation and
  correspondence}.
\newblock In \emph{Proceedings of the 58th Annual Meeting of the Association
  for Computational Linguistics}, pages 7155--7165, Online. Association for
  Computational Linguistics.

\bibitem[{Snover et~al.(2006)Snover, Dorr, Schwartz, Micciulla, and
  Weischedel}]{Snover06astudy}
Matthew Snover, Bonnie Dorr, Richard Schwartz, Linnea Micciulla, and Ralph
  Weischedel. 2006.
\newblock \href {http://www.cs.umd.edu/~snover/pub/LAMP_126.pdf} {A study of
  translation error rate with targeted human annotation}.
\newblock In \emph{In Proceedings of the Association for Machine Transaltion in
  the Americas (AMTA 2006)}.

\bibitem[{Song et~al.(2018)Song, Zhang, Wang, and
  Gildea}]{song-etal-2018-graph}
Linfeng Song, Yue Zhang, Zhiguo Wang, and Daniel Gildea. 2018.
\newblock \href {https://doi.org/10.18653/v1/P18-1150} {A graph-to-sequence
  model for {AMR}-to-text generation}.
\newblock In \emph{Proceedings of the 56th Annual Meeting of the Association
  for Computational Linguistics (Volume 1: Long Papers)}, pages 1616--1626,
  Melbourne, Australia. Association for Computational Linguistics.

\bibitem[{Trisedya et~al.(2018)Trisedya, Qi, Zhang, and
  Wang}]{trisedya-etal-2018-gtr}
Bayu~Distiawan Trisedya, Jianzhong Qi, Rui Zhang, and Wei Wang. 2018.
\newblock \href {https://doi.org/10.18653/v1/P18-1151} {{GTR}-{LSTM}: A triple
  encoder for sentence generation from {RDF} data}.
\newblock In \emph{Proceedings of the 56th Annual Meeting of the Association
  for Computational Linguistics (Volume 1: Long Papers)}, pages 1627--1637,
  Melbourne, Australia. Association for Computational Linguistics.

\bibitem[{Wang et~al.(2018)Wang, Pan, Huang, Zhang, Jiang, Ji, and
  Knight}]{wang-etal-2018-describing}
Qingyun Wang, Xiaoman Pan, Lifu Huang, Boliang Zhang, Zhiying Jiang, Heng Ji,
  and Kevin Knight. 2018.
\newblock \href {https://doi.org/10.18653/v1/W18-6502} {Describing a knowledge
  base}.
\newblock In \emph{Proceedings of the 11th International Conference on Natural
  Language Generation}, pages 10--21, Tilburg University, The Netherlands.
  Association for Computational Linguistics.

\bibitem[{Wang et~al.(2020{\natexlab{a}})Wang, Wan, and
  Jin}]{wang-etal-2020-amr}
Tianming Wang, Xiaojun Wan, and Hanqi Jin. 2020{\natexlab{a}}.
\newblock \href {https://doi.org/10.1162/tacl_a_00297} {{AMR}-to-text
  generation with graph transformer}.
\newblock \emph{Transactions of the Association for Computational Linguistics},
  8:19--33.

\bibitem[{Wang et~al.(2020{\natexlab{b}})Wang, Wang, An, Yu, and
  Chen}]{wang-etal-2020-towards}
Zhenyi Wang, Xiaoyang Wang, Bang An, Dong Yu, and Changyou Chen.
  2020{\natexlab{b}}.
\newblock \href {https://doi.org/10.18653/v1/2020.acl-main.101} {Towards
  faithful neural table-to-text generation with content-matching constraints}.
\newblock In \emph{Proceedings of the 58th Annual Meeting of the Association
  for Computational Linguistics}, pages 1072--1086, Online. Association for
  Computational Linguistics.

\bibitem[{Wiseman et~al.(2017)Wiseman, Shieber, and
  Rush}]{wiseman-etal-2017-challenges}
Sam Wiseman, Stuart Shieber, and Alexander Rush. 2017.
\newblock \href {https://doi.org/10.18653/v1/D17-1239} {Challenges in
  data-to-document generation}.
\newblock In \emph{Proceedings of the 2017 Conference on Empirical Methods in
  Natural Language Processing}, pages 2253--2263, Copenhagen, Denmark.
  Association for Computational Linguistics.

\bibitem[{Wolf et~al.(2020)Wolf, Debut, Sanh, Chaumond, Delangue, Moi, Cistac,
  Rault, Louf, Funtowicz, Davison, Shleifer, von Platen, Ma, Jernite, Plu, Xu,
  Le~Scao, Gugger, Drame, Lhoest, and Rush}]{wolf-etal-2020-transformers}
Thomas Wolf, Lysandre Debut, Victor Sanh, Julien Chaumond, Clement Delangue,
  Anthony Moi, Pierric Cistac, Tim Rault, Remi Louf, Morgan Funtowicz, Joe
  Davison, Sam Shleifer, Patrick von Platen, Clara Ma, Yacine Jernite, Julien
  Plu, Canwen Xu, Teven Le~Scao, Sylvain Gugger, Mariama Drame, Quentin Lhoest,
  and Alexander Rush. 2020.
\newblock \href {https://doi.org/10.18653/v1/2020.emnlp-demos.6} {Transformers:
  State-of-the-art natural language processing}.
\newblock In \emph{Proceedings of the 2020 Conference on Empirical Methods in
  Natural Language Processing: System Demonstrations}, pages 38--45, Online.
  Association for Computational Linguistics.

\bibitem[{Zhang* et~al.(2020)Zhang*, Kishore*, Wu*, Weinberger, and
  Artzi}]{Zhang*2020BERTScore:}
Tianyi Zhang*, Varsha Kishore*, Felix Wu*, Kilian~Q. Weinberger, and Yoav
  Artzi. 2020.
\newblock \href {https://openreview.net/forum?id=SkeHuCVFDr} {Bertscore:
  Evaluating text generation with bert}.
\newblock In \emph{International Conference on Learning Representations}.

\bibitem[{Zhao et~al.(2020)Zhao, Walker, and
  Chaturvedi}]{zhao-etal-2020-bridging}
Chao Zhao, Marilyn Walker, and Snigdha Chaturvedi. 2020.
\newblock \href {https://doi.org/10.18653/v1/2020.acl-main.224} {Bridging the
  structural gap between encoding and decoding for data-to-text generation}.
\newblock In \emph{Proceedings of the 58th Annual Meeting of the Association
  for Computational Linguistics}, pages 2481--2491, Online. Association for
  Computational Linguistics.

\end{thebibliography}
\bibliographystyle{acl_natbib}

\appendix
\section{Hyperparameters and Statistics of the Model}
\begin{table}[!htb]
\centering
\small
\begin{tabularx}{\linewidth}{>{\hsize=1.6\hsize}X>{\centering\arraybackslash\hsize=0.7\hsize}X>{\centering\arraybackslash\hsize=0.7\hsize}X}
\toprule
&\textbf{Origin}\footnotemark  &\textbf{+ Position}\\
\midrule
BART-base&139.42M&139.43M\\
distil-BART-xsum&305.51M&305.53M\\
BART-large&406.29M&406.31M\\
T5-base&222.88M&222,90M\\
T5-large&737.64M&737.65M\\
\bottomrule
\end{tabularx}
\caption{\# of Model Parameters \label{tab:para}}
\end{table}
 Our model is built based on the Huggingface framework \cite{wolf-etal-2020-transformers}\footnote{https://github.com/huggingface/transformers}. Because the average lengths for source and target text in the training set are 31 and 22 words respectively, we set the maximum length for both source and target to 100 words. For T5 preprocessing, we prepend ``translate RDF to English:'' before the input. For BART-base, distil-BART-xsum, and T5-base, we use a batch size of 32 and train the model. We use a batch size of 16 for Bart-large, and 6 for T5-large. We use the Adam optimizer \cite{kingma2014adam} to optimize each model with learning rate of $3\times10^{-5}$ with $\epsilon=1\times10^{-8}$ for a maximum of 10 epochs. We run each experiment on one Nvidia Tesla V100 GPU with 16G DRAM. We first fine-tuned the PLMs on crawled Wikipedia pairs for 3 epochs. The Wikipedia Fine-tuning stage takes about 24 hours for T5-large and 10 hours for the rest of models. The final WebNLG fine-tuning stage takes less than 1 hour for all the models. We chose our best model based on multi-BLEU score\footnote{https://gitlab.com/webnlg/webnlg-baseline/-/blob/master/multi-bleu.perl}. For inference, we use beam search with beam size in the range \{3,5\}. Table \ref{tab:para} shows the number of the parameters for each pre-trained model.

\footnotetext{\# of parameters are slightly different because we add special tokens to the vocabulary} 
\onecolumn
\section{Sample Generation Results}

\begin{table*}[ht!]
\small
 \begin{tabularx}{\linewidth}{>{\hsize=.2\hsize}X>{\hsize=1.8\hsize}X}
\toprule
\textbf{Category}&\textbf{Output}\\
\midrule

Input& S$|$ Aaron Turner P$|$ associatedBand/associatedMusicalArtist O$|$ \colorbox{fl9!50}{Twilight (band)} \newline
S$|$ Aaron Turner P$|$ associatedBand/associatedMusicalArtist O$|$ \colorbox{fl4!50}{Lotus Eaters (band)} \newline
S$|$ Aaron Turner P$|$ genre O$|$ \colorbox{fl5!50}{Sludge metal} \newline
S$|$ Aaron Turner P$|$ origin O$|$ \colorbox{fl6!50}{Massachusetts} \newline
S$|$ Aaron Turner P$|$ activeYearsStartYear O$|$ \colorbox{fl7!50}{1995}

\\\hdashline
Reference&
Aaron Turner was born in \colorbox{fl6!50}{Massachusetts} and started performing in \colorbox{fl7!50}{1995}. He formerly played with the band \colorbox{fl9!50}{Twilight} and is now in the \colorbox{fl4!50}{Lotus Eaters}. He is is closely associated with the \colorbox{fl5!50}{Sludge metal} genre.
\\\hdashline
T5-large&

Aaron Turner, who is from \colorbox{fl6!50}{Massachusetts}, is a member of the \colorbox{fl9!50}{Twilight} band and the \colorbox{fl4!50}{Lotus Eaters} band. He is also associated with the \colorbox{fl5!50}{Sludge metal} genre and started playing in \colorbox{fl7!50}{1995}.
\\\hdashline
T5-large \newline+ Wiki& 

Aaron Turner was born in \colorbox{fl6!50}{Massachusetts} and started playing in \colorbox{fl7!50}{1995}. He is associated with the band \colorbox{fl9!50}{Twilight} and the band \colorbox{fl4!50}{Lotus Eaters}. His music style is \colorbox{fl5!50}{Sludge metal}. 
\\\hdashline
T5-large +\newline Position &
Aaron Turner, who was born in \colorbox{fl6!50}{Massachusetts}, is associated with \colorbox{fl9!50}{Twilight} and \colorbox{fl4!50}{Lotus Eaters}. He is a member of \colorbox{fl5!50}{Sludge metal} band and has also been a musical artist. \textcolor{ao(english)}{[1995]}

\\\hdashline
T5-large \newline+ Wiki + \newline Position & 
Aaron Turner was born in \colorbox{fl6!50}{Massachusetts} and started playing \colorbox{fl5!50}{Sludge metal} in \colorbox{fl7!50}{1995}. He is also associated with \colorbox{fl9!50}{Twilight} and \colorbox{fl4!50}{Lotus Eaters}.
\\
\bottomrule
\end{tabularx}
\caption{Human and System Generated Description  We use the color box to frame each entity out with the same color as the corresponding entity. We highlight \textcolor{bleudefrance}{\textit{fabricate facts}}, \textcolor{ao(english)}{[missed relations]}, and \textcolor{auburn}{incorrect relations} with different color. }
\end{table*}

\begin{table*}[ht!]
\small
 \begin{tabularx}{\linewidth}{>{\hsize=.2\hsize}X>{\hsize=1.8\hsize}X}
\toprule
\textbf{Category}&\textbf{Output}\\
\midrule

Input& 
S$|$ American Civil War P$|$ commander O$|$ \colorbox{fl9!50}{Jefferson Davis} \newline
S$|$ Alfred Moore Scales P$|$ country O$|$ \colorbox{fl4!50}{United States} \newline
S$|$ United States P$|$ ethnicGroup O$|$ \colorbox{fl5!50}{Native Americans in the United States} \newline
S$|$ Alfred Moore Scales P$|$ battles O$|$ \colorbox{fl6!50}{Siege of Petersburg} \newline
S$|$ Siege of Petersburg P$|$ isPartOfMilitaryConflict O$|$ \colorbox{fl7!50}{American Civil War}
target
\\\hdashline
Reference&
\colorbox{fl7!50}{American Civil War}, \colorbox{fl6!50}{Siege of Petersburg} involved \colorbox{fl4!50}{U.S.} born Alfred Moore Scales. The American Civil War was commanded by \colorbox{fl9!50}{Jefferson Davis}. An ethnic group of the United States includes \colorbox{fl5!50}{Native Americans}.
\\\hdashline
T5-large&
Alfred Moore Scales is from the \colorbox{fl4!50}{United States} where \colorbox{fl5!50}{Native Americans} are one of the ethnic groups. He fought in the \colorbox{fl6!50}{Siege of Petersburg} which was part of the \colorbox{fl7!50}{American Civil War} commanded by \colorbox{fl9!50}{Jefferson Davis}.
\\\hdashline
T5-large \newline+ Wiki& 
\textcolor{auburn}{\textbf{The American Civil War was fought in the \colorbox{fl6!50}{Siege of Petersburg}.}} \colorbox{fl9!50}{Jefferson Davis} was the commander of the war. Alfred Moore Scales was born in the \colorbox{fl4!50}{United States} where \colorbox{fl5!50}{Native Americans} are an ethnic group. \textcolor{ao(english)}{[S$|$ Alfred Moore Scales P$|$ battles O$|$ Siege of Petersburg]}
\\\hdashline
T5-large +\newline Position &
Alfred Moore Scales was born in the \colorbox{fl4!50}{United States}, where \colorbox{fl5!50}{Native Americans} are an ethnic group. He fought in the American Civil War, which was led by \colorbox{fl9!50}{Jefferson Davis}. The \colorbox{fl6!50}{Siege of Petersburg} is part of the \colorbox{fl7!50}{American Civil War}. \textcolor{ao(english)}{[S$|$ Alfred Moore Scales P$|$ battles O$|$ Siege of Petersburg]}
\\\hdashline
T5-large \newline+ Wiki + \newline Position & 
Alfred Moore Scales is from the \colorbox{fl4!50}{United States} where \colorbox{fl5!50}{Native Americans} are one of the ethnic groups. He fought in the \colorbox{fl6!50}{Siege of Petersburg} which is part of the \colorbox{fl7!50}{American Civil War}. \colorbox{fl9!50}{Jefferson Davis} was the commander of the American Civil War.
\\
\bottomrule
\end{tabularx}
\caption{Human and System Generated Description  We use the color box to frame each entity out with the same color as the corresponding entity. We highlight \textcolor{bleudefrance}{\textit{fabricate facts}}, \textcolor{ao(english)}{[missed relations]}, and \textcolor{auburn}{\textbf{incorrect relations}} with different color. }
\end{table*}

\begin{table*}[ht!]
\small
\begin{tabularx}{\linewidth}{>{\hsize=.2\hsize}X>{\hsize=1.8\hsize}X}
\toprule
\textbf{Category}&\textbf{Output}\\
\midrule

Input& S$|$ \colorbox{fl9!50}{101 Helena} P$|$ discoverer O$|$ James Craig Watson \newline
S$|$ James Craig Watson P$|$ almaMater O$|$ \colorbox{fl4!50}{University of Michigan} \newline
S$|$ 101 Helena P$|$ discovered O$|$ \colorbox{fl5!50}{1868-08-15} \newline
S$|$ James Craig Watson P$|$ nationality O$|$ \colorbox{fl6!50}{Canada}\newline
S$|$ James Craig Watson P$|$ deathPlace O$|$ \colorbox{fl7!50}{Madison, Wisconsin}
\\\hdashline
Reference&
James Craig Watson, who discovered \colorbox{fl9!50}{101 Helena} on \colorbox{fl5!50}{15th August 1868}, is a \colorbox{fl6!50}{Canadian} national who attended the \colorbox{fl4!50}{University of Michigan}. He died in \colorbox{fl7!50}{Madison, Wisconsin}.
\\\hdashline
T5-large&
James Craig Watson is a \colorbox{fl6!50}{Canadian} who graduated from the \colorbox{fl4!50}{University of Michigan}. He was the discoverer of \colorbox{fl9!50}{101 Helena} which was discovered on \colorbox{fl5!50}{15 August 1868}. He died in \colorbox{fl7!50}{Madison, Wisconsin}.
\\\hdashline
T5-large \newline+ Wiki& 
James Craig Watson, a \colorbox{fl6!50}{Canadian}, graduated from the \colorbox{fl4!50}{University of Michigan} and discovered \colorbox{fl9!50}{101 Helena} on \colorbox{fl5!50}{August 15th, 1868}. He died in \colorbox{fl7!50}{Madison, Wisconsin}.
\\\hdashline
T5-large +\newline Position &
James Craig Watson, a \colorbox{fl6!50}{Canadian} , graduated from the \colorbox{fl4!50}{University of Michigan} and was the discoverer of \colorbox{fl9!50}{101 Helena} on \colorbox{fl5!50}{August 15, 1868}. He died in \colorbox{fl7!50}{Madison, Wisconsin}.
\\\hdashline
T5-large \newline+ Wiki + \newline Position & 
James Craig Watson, a \colorbox{fl6!50}{Canadian}, graduated from the \colorbox{fl4!50}{University of Michigan} and discovered \colorbox{fl9!50}{101 Helena} on \colorbox{fl5!50}{August 15th, 1868}. He died in \colorbox{fl7!50}{Madison, Wisconsin}.
\\
\bottomrule
\end{tabularx}
\caption{Human and System Generated Description  We use the color box to frame each entity out with the same color as the corresponding entity. We highlight \textcolor{bleudefrance}{\textit{fabricate facts}}, \textcolor{ao(english)}{[missed relations]}, and \textcolor{auburn}{\textbf{incorrect relations}} with different color. }
\end{table*}

\end{document}